%% file: main.tex
\pdfoutput=1
\documentclass[10pt,twocolumn,letterpaper]{article}
\usepackage[pagenumbers]{cvpr}
\usepackage[T1]{fontenc}
\usepackage{tabularx}
\usepackage{array}
\usepackage{pifont}
\definecolor{cvprblue}{rgb}{0.21,0.49,0.74}
\usepackage[pagebackref,breaklinks,colorlinks,allcolors=cvprblue]{hyperref}
\newcolumntype{L}{>{\raggedright\arraybackslash}X}
\newcolumntype{P}[1]{>{\raggedright\arraybackslash}p{#1}}

\definecolor{ours}{HTML}{E5186A}
\newcommand{\method}{STAT\xspace}

\definecolor{wxpun}{HTML}{BDBDBD}\definecolor{wxkey}{HTML}{8C8C8C}\definecolor{wxink}{HTML}{1A1A1A}
\definecolor{wxyes}{HTML}{029992}\definecolor{wxno}{HTML}{91571B}\definecolor{wxloc}{HTML}{2D4C9C}
\DeclareMathSizes{7.5}{7}{5}{5}%
\newcommand{\wxp}[1]{\textcolor{wxpun}{#1}}
\newcommand{\wxs}[2]{\wxp{"}\textcolor{#1}{#2}\wxp{"}}
\newcommand{\wxb}[2]{\wxp{"}\textbf{\textcolor{#1}{#2}}\wxp{"}}
\newcommand{\wxk}[1]{\wxp{"}\textcolor{wxkey}{#1}\wxp{":} }
\newcommand{\wxn}[1]{\wxb{wxink}{#1}\wxp{: [}}
\newcommand{\wxf}[4]{\wxp{\{}\wxk{finding}\wxs{wxink}{#1}\wxp{,} & \wxk{present}\wxb{wx#2}{#2}\wxp{,} & \wxk{location}\wxs{wxloc}{#3}\wxp{,} & \wxk{detail}#4}
\newcommand{\wxnull}{\textcolor{wxpun}{null}}

\title{Clinical Reports Are Not Image Captions}
\author{%
Ömer Faruk Durugöl$^{1}$ \quad Jiayi Wang$^{1}$ \quad Teodora Spasojevic$^{1}$ \quad Hadrien Reynaud$^{1}$\\
Ibrahim Ethem Hamamci$^{2,3,4}$ \quad Sezgin Er$^{2,3,4}$ \quad Suprosanna Shit$^{3}$\\
Utku Turkbey$^{2}$ \quad Bjoern Menze$^{3}$ \quad Bernhard Kainz$^{1,5}$\\[0.4em]
{\normalsize $^{1}$Friedrich-Alexander-Universität Erlangen-Nürnberg \quad $^{2}$Forithmus \quad $^{3}$University of Zurich}\\
{\normalsize $^{4}$Istanbul Medipol University \quad $^{5}$Imperial College London}%
}

\begin{document}

\twocolumn[{%
\renewcommand\twocolumn[1][]{#1}%
\maketitle
\vspace{-2.6em}
\begin{center}
\includegraphics{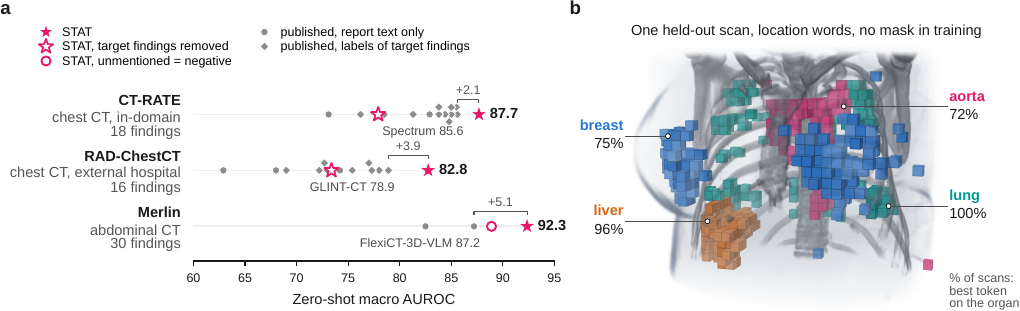}
\captionof{figure}{(a) Zero-shot macro AUROC of published 3D CT models (gray; diamonds train on labels of the target findings, as \method does, circles on report text only) and of \method (star) on CT-RATE (18 findings), RAD-ChestCT from an external hospital, never seen in training (16 findings), and Merlin abdominal CT (30 findings). Brackets give the lead over the prior state-of-the-art zero-shot model. Open stars show \method trained with the target findings removed (unseen classes), and the open circle scores unmentioned Merlin findings as negatives instead of skipping them. fVLM and CT-DiagVLM report 16 CT-RATE and 14 RAD-ChestCT findings.
(b) Four location words queried on one held-out chest CT. For each word, the 1\% of the body's 12\,mm cubes of patch tokens (8 slices of 1.5\,mm) with the highest cosine are drawn see-through. Percentages give the share of held-out scans whose best-scoring token lies on the organ's automatic CADS mask~\cite{cads} (chance: liver 5\%, breast 5\%, lung 16\%, aorta 1\%). Over 24 location words, the mean is 58\% (chance 2.5\%).}
\label{fig:teaser}
\end{center}
\vspace{0.6em}
}]

\begin{abstract}
Vision-language models for 3D computed tomography (CT) learn from scans paired with radiology reports, yet most treat the report, whole or in pieces, as a caption. A report is closer to a checklist: it states dozens of findings, many of them negative, each in a small part of the scan, and often says where. We propose \textbf{STAT} (\textbf{ST}atus-\textbf{A}ligned \textbf{T}yped findings), which supervises every reported finding, with its status and its stated location, on the patch tokens of the scan. An open-weight language model rewrites each report sentence into typed findings: a finding name, a status of present, absent, or uncertain, and a location. The status sets the target, so a negated mention never trains its finding as present. Each finding is scored against all patch tokens by log-sum-exp pooling, and its lobe or side is predicted at the tokens it selects, without masks or boxes. Only a linear head of 0.65M parameters is trained between frozen encoders, so one design choice can change at a time. On these encoders, report sentences add about 20 points of area under the receiver operating characteristic curve (AUROC) over whole reports in-domain, typed findings add 4 more, more than half of it through the status, and patch tokens and the CT encoder (over a natural-image one) add 2 to 6 each. Trained on these weak labels, which include the benchmark findings, STAT reaches the highest zero-shot macro AUROC on in-domain chest CT (87.7), chest CT from an external hospital (82.8), and abdominal CT (92.3), ahead of models that train their encoders. It outperforms prior zero-shot models, which also train on labels of the benchmark findings, by 2.1 points in-domain and 3.9 at the external hospital, where the prior state-of-the-art zero-shot model scores 78.9 and the classifier 79.6. The same head points at findings and organs without mask supervision, and training stated locations as words lifts organ words from 42\% to 58\% on their organ.%
\end{abstract}

\section{Introduction}
\label{sec:intro}

Supervision decides what a vision model can recognize. A classifier recognizes only its label set~\cite{deng2009imagenet}. CLIP~\cite{clip} replaced the label set with web captions and turned any sentence into a classifier, and 3D computed tomography (CT) models now train this way on scans paired with radiology reports~\cite{ctclip,merlin,fvlm,colipri,flexict}. Their main test is zero-shot detection of findings such as pleural effusion or lung nodules, scored from text alone with no downstream classifier fitted on the benchmark's labels.

A radiology report differs from a caption on both sides of the pair. On the text side, it states dozens of findings about one volume, many of them negative (``No pleural effusion.''), often with their location (``in the right upper lobe'')~\cite{jain2021radgraph}, and much of it is boilerplate. 59\% of CT-RATE validation sentences appear word for word in at least 50 training reports. On the image side, the evidence for a finding is small. At 16-pixel patches, a CT-RATE scan gives about 180k tokens, and a small nodule touches only a few of them. CLIP pools each side into one vector, so the signal of one abnormal finding has to survive an average over the report and another over the scan.

Recent CT models depart from whole-report CLIP in four ways. They scale encoders and data~\cite{colipri,glintct,flexict,spectre,zhao2026towards}, split the report or the scan by anatomy~\cite{merlin,fvlm,arcct,jolia,a2moe}, train on findings that a language model extracts from each report~\cite{mpsct,greenrfm,radfinder}, or align report text with local image features~\cite{glintct,radzero3d,li2025medvista3d}. Together these changes raised the zero-shot macro-averaged area under the receiver operating characteristic curve (AUROC) on CT-RATE from 73.1~\cite{ctclip} to 85.6~\cite{spectrum}. Each work changes the encoder, the data, the unit of text, the matching, and the loss at once, and the published zero-shot models in \cref{tab:main} lose 5.5 to 12.9 points at an external hospital (RAD-ChestCT~\cite{radchest}). Which of these changes matters, and which of them survives a change of hospital, is an open question.

\begin{figure*}[tp]
\centering
\includegraphics{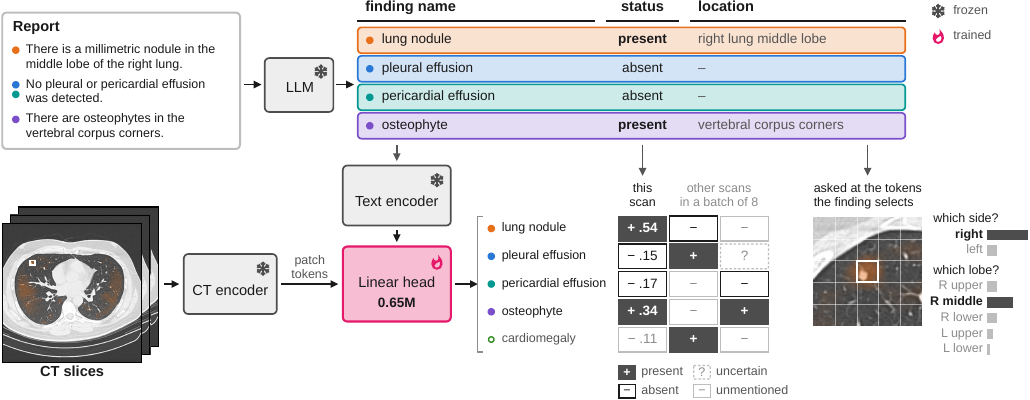}
\caption{Training, shown on a held-out scan and its report, with the trained head's scores.
A language model rewrites each report sentence into typed findings, each a finding name with its status and its location. Frozen encoders embed the names and the patch tokens of the scan, and a linear head scores every finding against every token. Log-sum-exp pooling over tokens gives the finding score, which is trained with a sigmoid loss against the target that the status sets. The head asks a stated side or lung lobe at the tokens that the finding itself selects. The per-token scores form the query map.}
\label{fig:method}
\end{figure*}

To answer this question, we freeze both encoders, a CT encoder pretrained without reports on 266k CT volumes and an 8B text embedding model, so that one design choice can change at a time, and train only a linear head of 0.65M parameters, with the same loss and settings on chest and abdominal CT. We present \textbf{\method} (\textbf{ST}atus-\textbf{A}ligned \textbf{T}yped findings), which supervises each reported finding, with its status and its stated location, on the patch tokens of the scan (\cref{fig:method}). An open-weight language model rewrites each report sentence into typed findings: a finding name, a status of present, absent, or uncertain, and a location. No label is written by hand, so these targets are weak labels~\cite{weaksup}. The status sets the target of each finding: only present mentions are positives, so a negated mention never trains its finding as present. Each finding is scored against all patch tokens through log-sum-exp pooling, so that a finding in a few tokens survives. Where the report states a location, the head asks it at the tokens that the finding itself selects, with no mask, box, or anatomy network. Prior CT models use extracted findings as further captions~\cite{colipri,flexict} or as labels on one pooled vector~\cite{mpsct,greenrfm,radfinder} and train the image encoder, whereas \method uses them as per-finding targets on the patch tokens of frozen encoders and adds the stated location.

Trained on these weak labels, this linear head has the highest zero-shot macro AUROC on CT-RATE, on RAD-ChestCT from an external hospital, and on Merlin abdominal CT, ahead of models that train their encoders (\cref{fig:teaser}a). These weak labels include the benchmark findings, as do the training labels of eleven of the 14 zero-shot chest CT models in \cref{tab:main}. RAD-ChestCT contributes no scan and no report to training. With every target finding removed from the training text, \method scores 73.4 at the external hospital, above six of the 13 published zero-shot models with a RAD-ChestCT score. At the external hospital and on Merlin, \method also scores above every published classifier trained on labels of the target findings. The controlled comparisons show why (\cref{fig:ablations}). On frozen encoders, the unit of supervision is the main lever. Report sentences, which prior work already uses, add about 20 points over whole reports on CT-RATE but stay below the prior state-of-the-art zero-shot models, and typed findings add 4.4 more and pass them, a step comparable to that of patch tokens or of the CT encoder. More than half of it comes from the status: without it, a negated mention such as ``no effusion'' trains effusion as present. The per-token scores point at findings without masks. The stated location adds little to classification and acts on where these scores point, and location words such as ``liver'' find their organs (\cref{fig:teaser}b).
Our contributions are:
\begin{itemize}
\setlength{\itemsep}{0pt}
\item Typed findings on tokens. Every reported finding supervises the patch tokens of frozen encoders, and its status sets the target. To our knowledge, \method is the first report-supervised model to ask a report-stated location at a finding's own tokens, without masks, boxes, detectors, or anatomy networks.
\item Controlled comparisons. One design choice at a time on frozen encoders (unit of text, status, location, matching, image encoder, text encoder), each tested in-domain and at an external hospital.
\item Localization from reports. The same head, tested against radiologist masks, organ masks, and the locations and sizes that reports state.
\end{itemize}

\section{Related work}
\label{sec:related}

\noindent\textbf{Image-text learning.} CLIP~\cite{clip} learns from web image-text pairs, and SigLIP~\cite{siglip} replaces the softmax with a sigmoid per pair, as our loss does. LiT~\cite{lit} shows that a frozen image tower suffices for zero-shot transfer when the text tower is trained. FILIP~\cite{filip} and ColBERT~\cite{colbert} match tokens instead of pooled vectors. CLIP-style models largely ignore word order~\cite{negclip} and negation~\cite{negbench}. \method freezes both towers, matches each finding to the image tokens, moves negation from the text into the targets, and asks the stated location at the finding's tokens.

\noindent\textbf{Report-derived labels and 2D medical models.} Rule-based labelers mine findings from reports together with their negation~\cite{chapman2001simple} or uncertainty, and the labeler of CheXpert~\cite{irvin2019chexpert} fixes 14 observations before training. ChestX-ray8~\cite{chestxray8} localizes findings from such weak labels with log-sum-exp pooling~\cite{pinheiro2015image}. ConVIRT~\cite{convirt} brought image-report contrastive learning to chest X-rays, GLoRIA~\cite{gloria} and MGCA~\cite{mgca} add local alignment, and CheXzero~\cite{chexzero} contrasts a positive with a negative prompt. MedKLIP~\cite{medklip} and KAD~\cite{kad} extract entities with their existence and query the image with them, and MedKLIP also predicts the position of each entity from the pooled output of its query, with no loss on the image patches. RadZero~\cite{radzero} matches finding sentences to the patches of a frozen X-ray encoder. \method extends the entity and status supervision of MedKLIP and KAD to an open vocabulary of 15k concepts on the patch tokens of 3D scans, and asks the position at the tokens of the finding.

\noindent\textbf{3D CT vision-language models.} CT-CLIP~\cite{ctclip} released CT-RATE and trained whole-report CLIP on chest CT. Merlin~\cite{merlin} splits abdominal reports into sections and adds diagnosis codes. fVLM~\cite{fvlm}, ARC-CT~\cite{arcct}, and SCOPE~\cite{scope} align organ regions from segmentation masks with organ-level text, and Jolia~\cite{jolia} learns one image query per organ section without masks. COLIPRI~\cite{colipri} and FlexiCT-3D-VLM~\cite{flexict} obtain positive and negative findings from a language model and contrast ``X.'' with ``No X.'' Li~\emph{et~al.}~\cite{mpsct}, GreenRFM~\cite{greenrfm}, and RadFinder~\cite{radfinder} train on labels that a language model extracts from the reports, and CorLIP~\cite{corlip} treats a text as positive whenever its clinical assertions agree with the image. GLINT-CT~\cite{glintct} and RadZero3D~\cite{radzero3d} match one-finding sentences from a language model to patch tokens, and GLINT-CT segments findings zero-shot. All of these models train their image encoder, fully or in part. COLIPRI and FlexiCT-3D-VLM use extracted findings as text samples for one pooled vector, whereas \method uses them as per-finding targets on the tokens of frozen encoders and takes locations from the report, not from masks.

\section{Method}
\label{sec:method}

In our approach, a report supervises a scan through the findings it states, each with its status and its location (\cref{fig:method}). Only a linear head between two frozen encoders is trained.

\noindent\textbf{Typed findings.} A report $r$ consists of sentences $u_1,\dots,u_M$. A language model $C$, the compiler, maps each sentence to the set of findings it states, $C(u_m)=\{(n,z,\ell,\delta)\}$, with a finding name $n$, a status $z\in\{\text{present},\text{absent},\text{uncertain}\}$, a location $\ell$, and a detail $\delta$ such as size or number. For example, ``No effusion. A 4\,mm nodule in the right upper lobe.'' becomes (pleural effusion, absent) and (lung nodule, present, right upper lobe, 4\,mm). Training uses the name, the status, and the location, and the detail serves a size test in \cref{sec:experiments}. Let $\mathcal{N}$ be the set of names over all compiled reports and $F$ a frozen text encoder. Names are merged into concepts by single linkage: two names share a concept when $\cos(F(n),F(n'))\geq\gamma$, unless one carries a word of increase (``enlarged'') or of decrease (``reduced'') that the other lacks. The resulting map $\mu:\mathcal{N}\rightarrow\mathcal{C}$ sends each name to its concept, and each concept is named by its most frequent member. For a training scan $x$ with report $r$, let $Z_{xc}$ be the multiset of statuses with which $r$ mentions concept $c$. The scan-level status $z_{xc}$ is present if present $\in Z_{xc}$, absent if $Z_{xc}$ holds only absent, uncertain for any other nonempty $Z_{xc}$, and unmentioned ($\varnothing$) if the report leaves $c$ out. One present mention decides.

\noindent\textbf{Scoring on tokens.} A frozen image encoder $E$ maps a scan $x$ to tokens $v_1,\dots,v_N\in\mathbb{R}^{D_v}$, and $F$ maps the name of concept $c$ to $e_c\in\mathbb{R}^{D_t}$. For a 2D encoder applied slice by slice, the tokens of all slices form one set. The head projects both sides into a shared space of dimension $d$,
\begin{equation}
h_i=W_v\,\mathrm{LN}(v_i),\qquad q_c=W_t\,\mathrm{LN}(e_c),
\end{equation}
with LayerNorms~\cite{ba2016layer} and linear maps $W_v\in\mathbb{R}^{d\times D_v}$ and $W_t\in\mathbb{R}^{d\times D_t}$, which are the only trained layers. Because both encoders are frozen, tokens and text embeddings are computed once and cached. The score of concept $c$ in scan $x$ pools the cosine similarities $a_i=\cos(h_i,q_c)$ over all tokens with a log-sum-exp, a smooth maximum from multiple-instance learning~\cite{pinheiro2015image,chestxray8},
\begin{equation}
s(x,c) = \tau \Big( \log \sum_{i=1}^{N} \exp\big(a_i/\tau\big) - \log N \Big),
\label{eq:score}
\end{equation}
with a learned temperature $\tau$. The score moves from the mean of the $a_i$ towards their maximum as $\tau$ decreases, so a finding that a few tokens show keeps a high score. Its gradient with respect to $a_i$ is the pooling weight $\pi_i=\mathrm{softmax}_i(a/\tau)$, so training pushes most on the tokens that already match the finding. This is what lets the per-token cosines point at findings without masks.

\noindent\textbf{Status targets.} The status decides what a mention teaches. Each pair of a training scan $x$ and a concept $c$ receives a target $y_{xc}$ and a weight $w_{xc}$, with $(y,w)=(+1,1)$ for present, $(-1,1)$ for absent, $(-1,w_u)$ with $0<w_u<1$ for unmentioned ($\varnothing$), and $w=0$ for uncertain. Only present mentions enter the positives. The status has to act through the target, because $F$ does not separate a finding from its negation~\cite{negbench}: ``no X'' and ``X'' have a median cosine of .84, against .55 between different findings (\cref{sec:supp-details}). Absent mentions are full negatives. An unmentioned concept is a weak negative, because reports leave out many normal and irrelevant findings. It gets weight zero in a scan whose report states one of its near-synonyms as present (text cosine of at least .80), because reports name one finding in several ways. An uncertain mention is ignored, as in the U-Ignore policy of CheXpert~\cite{irvin2019chexpert}. For a batch $\mathcal{B}$, the loss sums over the concepts $\mathcal{C}_\mathcal{B}$, those mentioned in $\mathcal{B}$ and 256 further concepts per step, drawn in proportion to their training frequency,
\begin{equation}
\mathcal{L} = -\sum_{x\in\mathcal{B}}\sum_{c\in\mathcal{C}_\mathcal{B}} w_{xc} \log \sigma\big(y_{xc}\,(t\, s(x,c) + b)\big),
\end{equation}
a weighted sigmoid loss~\cite{siglip} with learned $t$ and $b$.

\noindent\textbf{Location.} A report often states where a finding lies, and this location enters training in two ways. First, 32 location words (five lung lobes, two sides, and 25 organs) are concepts of their own, present in a scan when a present abnormal finding is located there. Second, for a present finding $c$ whose location names a lung lobe or a side, the head asks this location at the tokens that the finding itself selects. Let $\mathcal{T}_c$ be the $K$ tokens with the highest $a_i$, $\pi_i$ their pooling weights, renormalized over $\mathcal{T}_c$, and $p_i(l)=\mathrm{softmax}_{l\in\mathcal{F}}(\cos(h_i,q_l)/\tau)$ the choice of token $i$ among the words of a family $\mathcal{F}$, the lobes or the sides. With $A_{xc}\subseteq\mathcal{F}$ the locations the report states,
\begin{equation}
\mathcal{L}_{\text{loc}} = -\sum_{x\in\mathcal{B}}\sum_{c}\sum_{i\in\mathcal{T}_c} \pi_i \log \sum_{l\in A_{xc}} p_i(l),
\label{eq:loc}
\end{equation}
and the head minimizes $\mathcal{L}+\lambda\,\mathcal{L}_{\text{loc}}$. The loss adds no parameter and needs no mask, box, or anatomy network.

At test time, a finding is queried by its name $n$ alone, such as ``pleural effusion'', with no prompt template, negative prompt, or mask. Its embedding $q_n=W_t\,\mathrm{LN}(F(n))$ gives the score $s(x,n)$ of \cref{eq:score} for classification and the query map $a_i=\cos(h_i,q_n)$ over tokens for localization. The best-scoring token $\arg\max_i a_i$ points at the finding, and the tokens with $a_i\geq\theta$ segment it.

\section{Experiments}
\label{sec:experiments}

\begin{table}[t]
\caption{Macro AUROC on the three benchmarks.
Upper block: zero-shot models, scored from a text query. Lower block: classifiers with one fixed output per target finding.
Labels: labels of the target findings in training, either none (report text only), LLM (extracted from reports by a language model), bench.\ (the benchmark's own labels), or removed (every target finding removed from the training text, \cref{sec:removed}).
Enc.: trains its image encoder (partly: adapters or some layers). RAD: RAD-ChestCT.
Published numbers use each paper's query protocol and test cohort. The chest scores of \method are means of three seeds.
$^\circ$16 and 14 findings, on which \method scores 88.3 and 83.3.
$^\dagger$Needs organ masks at test time.
$^\ddagger$Not trained on Merlin, prompt tuned on the Merlin test set.
$^\S$As evaluated by FlexiCT~\cite{flexict}. $^\P$As evaluated by Wang~\emph{et~al.}~\cite{wang2026dcppd}.
$^\star$22 of the 30 findings, on which \method scores 94.7.
$^\parallel$The image encoder saw the Merlin test images in image-only pretraining.
95\% intervals of \method: 87.0 to 88.2 and 82.3 to 83.3 (first seed), and 91.7 to 93.0.
}
\label{tab:main}
\centering
\footnotesize
\setlength{\tabcolsep}{0pt}
\begin{tabular*}{\columnwidth}{@{\extracolsep{\fill}}lccccc@{}}
\toprule
Model & Labels & Enc. & CT-RATE & RAD & Merlin \\
\midrule
CT-CLIP~\cite{ctclip} & none & yes & 73.1 & 62.9 & - \\
RadZero3D~\cite{radzero3d} & LLM & partly & 76.2 & 69.0 & - \\
fVLM$^{\dagger\circ}$~\cite{fvlm} & none & yes & 77.8 & 68.0 & - \\
COLIPRI~\cite{colipri} & LLM & yes & 78.5 & 73.0 & - \\
FlexiCT-3D-VLM~\cite{flexict} & LLM & yes & 81.3 & - & 87.2$^\parallel$ \\
BrgSA v2~\cite{brgsa} & none & yes & 82.9 & 74.2 & - \\
Li~\emph{et~al.}~\cite{mpsct} & LLM & yes & 83.8 & 77.3 & - \\
RadFinder~\cite{radfinder} & LLM & partly & 83.8 & 77.0 & - \\
CT-DiagVLM$^\circ$~\cite{ctdiagvlm} & bench. & yes & 84.4 & 75.4 & - \\
GreenRFM~\cite{greenrfm} & LLM & yes & 84.8 & 78.0 & - \\
GLINT-CT~\cite{glintct} & LLM & yes & 85.0 & 78.9 & - \\
SCOPE~\cite{scope} & bench. & yes & 85.0 & 72.2 & - \\
ARC-CT~\cite{arcct} & LLM & yes & 85.5 & 73.4 & - \\
Spectrum~\cite{spectrum} & bench. & yes & 85.6 & 72.7 & - \\
Merlin~\cite{merlin} & none & yes & - & - & 82.5$^\S$ \\
\midrule
\method (ours) & LLM & no & \textbf{87.7} & \textbf{82.8} & \textbf{92.3}$^\parallel$ \\
\quad with DINOv3 & LLM & no & 84.2 & 77.2 & - \\
\quad trained without target findings & removed & no & 77.9 & 73.4 & - \\
\quad unmentioned as negatives & LLM & no & - & - & 88.9$^\parallel$ \\
\midrule
\multicolumn{6}{@{}l}{\emph{Classifiers, one fixed output per target finding}} \\
Same head, CT-RATE's labels & bench. & no & 88.8 & 81.7 & - \\
AnyMC3D~\cite{anymc3d} & bench. & partly & 88.4 & - & - \\
Spectrum, linear probe~\cite{spectrum} & bench. & yes & 87.5 & 72.5 & - \\
NV-Reason-CT~\cite{nvreasonct} & LLM & yes & 87.1 & 79.6 & 83.3$^\ddagger$ \\
Pillar-0, linear$^\P$~\cite{pillar0} & LLM & no & 86.1 & 79.6 & - \\
GreenRFM, vision encoder~\cite{greenrfm} & LLM & yes & - & - & 87.4 \\
CT-IDP$^\star$~\cite{ctidp} & bench. & no & - & - & 89.7 \\
\bottomrule
\end{tabular*}
\end{table}

\noindent\textbf{Data.} Chest CT trains on CT-RATE~\cite{ctclip}: 45,850 volumes of 23,469 scans, one report per scan, after holding out the validation and test patients of ReXGroundingCT~\cite{rexgroundingct}. The in-domain test set is CT-RATE validation, 3,039 volumes of 1,304 patients with 18 findings. The external test set is RAD-ChestCT~\cite{radchest}, 3,630 volumes from an external hospital with 16 findings labeled from its reports. Abdominal CT trains on the 15,309 training studies of Merlin~\cite{merlin} and tests on its 5,125 test studies with 30 findings. Our image encoder saw all Merlin images, test studies included, in image-only pretraining, as did the encoder of the prior state-of-the-art zero-shot model on Merlin, FlexiCT-3D-VLM~\cite{flexict}, which belongs to the same family. All data are public and de-identified, so this retrospective study needed no further ethics approval.

\noindent\textbf{Implementation.} The image encoder $E$ is FlexiCT-2D~\cite{flexict} (0.14B parameters), a DINOv3-style vision transformer (ViT)~\cite{dosovitskiy2020image,dinov3} trained further on CT slices without reports. Its pretraining used 266,227 volumes from 56 public datasets, which include the CT-RATE training volumes and all Merlin volumes and exclude CT-RATE validation and RAD-ChestCT. It encodes each axial slice at 0.75\,mm in-plane with 16-pixel patches, so each token covers 12$\times$12\,mm. The text encoder $F$ is Qwen3-Embedding-8B~\cite{qwen3emb}, and the language model $C$ is Qwen3-30B-A3B~\cite{qwen3}, which reads the Findings and the Impression of each report. With $\gamma=.95$, the reports of CT-RATE give 14,630 concepts, 11 of them new location words. For Merlin, the prompt of $C$ names abdominal CT and a scan keeps 600\,mm instead of 360\,mm along the body axis, and nothing else changes. The head has $d=128$ and 0.65M parameters, $w_u=.3$, $K=128$, and $\lambda=.3$, and trains for 30k steps on every dataset. Compiling the reports takes under one GPU-hour and caching the tokens about 7, both once for every design, and the head then trains in about one hour on one H200 GPU. The head, the number of steps, and the learning rate were fixed without RAD-ChestCT or Merlin. The prompt, the concept inventory, and all settings are in \cref{sec:supp-compiler,sec:supp-concepts,sec:supp-details}. Code and model will be released.

\noindent\textbf{Protocol and metrics.} We call a result zero-shot when a model scores a finding from text alone, with no downstream classifier fitted on the benchmark's labels and no query tuned on its test set. This is the protocol of CLIP~\cite{clip} and of CT vision-language work~\cite{ctclip}, and it does not restrict the training signal: the weak labels of \method include the target findings unless we remove them (\cref{sec:removed}). On Merlin, we use the released test labels, which mark each finding as present, absent, or unmentioned. Each finding is scored present against absent, as in prior work~\cite{merlin,nvreasonct}, and studies that leave it unmentioned are skipped. We report macro AUROC on a 0 to 100 scale with 95\% bootstrap intervals over patients, and over studies on Merlin, whose release has no patient identifiers. \method has three seeds on chest CT (87.6 to 87.7 on CT-RATE, 82.7 to 82.8 on RAD-ChestCT), and we report their mean. The other configurations have one run unless stated, and differences use unrounded scores. CT-RATE served as our development set, so RAD-ChestCT, Merlin, and the datasets of \cref{tab:other} are the cleaner tests. One rule of the near-synonym list was added with the nodule scores of \cref{tab:other} in view (\cref{sec:supp-location}).

\begin{figure*}[tp]
\centering
\includegraphics{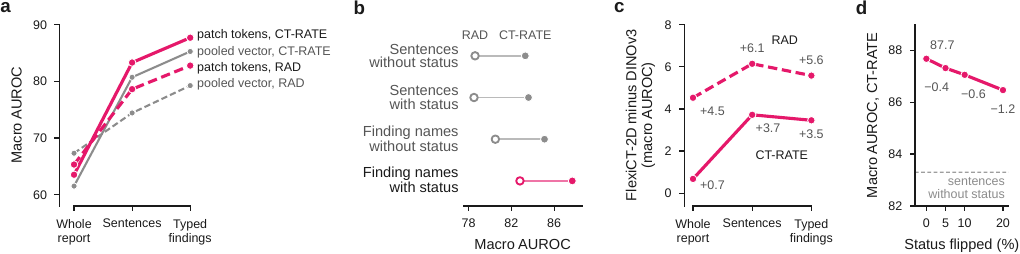}
\caption{(a) The unit of text is the main lever, and patch tokens (magenta) add about 2.5 points over a pooled vector (gray) on CT-RATE (solid) and about 4 on RAD-ChestCT (dashed), but lose 2.0 there with whole reports.
(b) What the status adds to sentences, by keeping only those that state a present finding, and to finding names (filled, CT-RATE, open, RAD-ChestCT).
(c) AUROC gain of the CT encoder (FlexiCT-2D) over the natural-image encoder (DINOv3) at each unit of text.
(d) Flipping the status of a fifth of the findings at random costs 1.2 points and keeps more than two thirds of the gain of typed findings over sentences (dashed, sentences without status).}
\label{fig:ablations}
\end{figure*}

\noindent\textbf{Baselines.} Every number of another model in \cref{tab:main} comes from a published paper. For controlled comparisons, the same head and loss express the designs of prior work as cells of a grid (\cref{fig:ablations}a): whole-report CLIP as in CT-CLIP (whole report, pooled vector); sentences on tokens, close to GLINT-CT and RadZero3D; extracted findings on one pooled vector as in Li~\emph{et~al.}, GreenRFM, and RadFinder; and \method (typed findings, patch tokens). All cells share the encoders, scans, number of steps, and learning rate, and neighboring cells differ in one choice (\cref{sec:supp-details} defines each cell). The whole-report cells and \method read Findings and Impression, the sentence cells Findings only; on Findings only, \method scores 87.6 and 82.8. The pooled cells replace the log-sum-exp by attention pooling with one learned query~\cite{ilse2018attention}. As a second image encoder we use DINOv3 ViT-B/16~\cite{dinov3}, which saw only natural images.

\subsection{Zero-shot classification}

Our proposed \method outperforms previous zero-shot models on the three main benchmarks (\cref{tab:main}). On chest CT, it improves over Spectrum~\cite{spectrum} on CT-RATE and GLINT-CT~\cite{glintct} on RAD-ChestCT, with a smaller performance drop on the external hospital dataset (RAD-ChestCT) than any of the compared models (\cref{fig:supp-transfer}). On abdominal CT, it also outperforms FlexiCT-3D-VLM, which trains on the same Merlin reports. These gains come from a small trained head, while both encoders remain frozen.

Further, we compare with classifiers trained directly on the target findings. AnyMC3D~\cite{anymc3d} slightly outperforms \method on CT-RATE, but \method outperforms the prior classifiers at the external RAD-ChestCT and on Merlin. On Merlin, GreenRFM~\cite{greenrfm} trains on report-derived labels, similar to our \method, while NV-Reason-CT~\cite{nvreasonct} has not trained on Merlin but tunes its prompt on the test set. Our proposed \method outperforms both. CT-IDP~\cite{ctidp} evaluates only 22 of the 30 Merlin findings; however, our \method also performs better on those same findings. For detailed training exposure, evaluation protocols, and finding coverage, please refer to \cref{tab:main}.

We also test \method on four additional public datasets covering COVID-19 detection, disease severity, and lung nodule malignancy (\cref{tab:other}). On MosMed~\cite{mosmed}, whose images were unseen by our encoder, \method improves over the prior state-of-the-art model that does not train on MosMed. Nevertheless, on nodule malignancy and disease severity, dedicated supervised models remain stronger (\cref{sec:supp-other}).

\noindent\textbf{Target findings removed from training.}\label{sec:removed}
Our main results follow the usual zero-shot protocol: a finding is queried by text, without fitting a downstream classifier. These results do not demonstrate generalization to entirely unseen findings~\cite{lampert2009learning}, because the report-derived training targets include the benchmark concepts, as they do for most competing methods. To measure this distinction, we remove every target finding, including its synonyms and subtypes, and retrain \method. We observe that performance falls by 9.8 points on CT-RATE and 9.4 on RAD-ChestCT (\cref{tab:main}), confirming that direct supervision on these concepts matters. Nevertheless, \method trained without the target findings still outperforms six of the thirteen published zero-shot models evaluated on RAD-ChestCT. Further, we remove the same amount of training information at random, which barely affects performance, confirming that the drop comes from removing the target findings specifically (\cref{sec:supp-strict}).

\begin{table}[tbp]
\caption{One design choice at a time with the same frozen encoders (macro AUROC).
RAD 16 and RAD 81 are the 16 standard findings of RAD-ChestCT and the 81 report-derived findings with at least 20 positives.
The reference rows are fitted on CT-RATE's labels.
Each row reports its best of three query forms on the test set (\cref{sec:supp-details}), a choice outside the zero-shot protocol that favors the other rows: for \method the best form is the label name, the single query of \cref{tab:main}. ``Name and status only'' trains without the location loss, the location words, and the near-synonym rule.}
\label{tab:ablation}
\centering
\footnotesize
\setlength{\tabcolsep}{0pt}
\begin{tabular*}{\columnwidth}{@{\extracolsep{\fill}}llccc@{}}
\toprule
Step & Text unit and matching & CT-RATE & RAD 16 & RAD 81 \\
\midrule
1 & Whole report, pooled vector & 61.5 & 67.3 & 64.8 \\
2 & Whole report, tokens & 63.5 & 65.3 & 63.3 \\
 & Sentences, pooled vector & 80.7 & 74.4 & 70.5 \\
3 & Sentences, tokens & 83.3 & 78.6 & 73.8 \\
 & Typed findings, pooled vector & 85.2 & 79.2 & 73.1 \\
4 & Typed findings, tokens (\method) & \textbf{87.7} & \textbf{82.8} & \textbf{76.8} \\
& \quad ``no X'' left out of training & 87.5 & 82.7 & 76.7 \\
 & \quad name and status only & 87.4 & 82.5 & 76.5 \\
  & \quad status flipped for 20\% & 86.5 & 81.9 & 75.9 \\
 & \quad names without status & 85.1 & 80.5 & 74.6 \\
  & \quad with Qwen3-Embedding-0.6B & 87.1 & 82.3 & 76.0 \\
   & \quad with CXR-BERT~\cite{cxrbert} & 84.4 & 80.1 & 73.5 \\
 & \quad with DINOv3 & 84.2 & 77.2 & 71.5 \\
\midrule
ref & Same head on CT-RATE's 18 labels & 88.8 & 81.7 & - \\
ref & Linear probe on the pooled vector & 83.8 & 71.4 & - \\
\bottomrule
\end{tabular*}
\end{table}

\subsection{What makes STAT work}

Our hypothesis is that the main improvement comes from \emph{what the model learns from a report}, rather than from a more complicated architecture (\cref{tab:ablation,fig:ablations}). We find that treating the whole report as one caption performs poorly. In contrast, splitting it into sentences makes a much larger difference, and replacing sentences with typed findings adds a further 4.4 points on CT-RATE and 4.2 on RAD-ChestCT, enough to pass the previous state-of-the-art zero-shot models. This observation holds both in-domain and on the external hospital dataset. Typed findings also keep their advantage over sentences when the text encoder is trained with LoRA~\cite{lora} (\cref{sec:supp-trained}). We further identify that matching findings to image patches rather than the entire scan also improves classification performance.

\noindent\textbf{Why do typed findings help?} The status gives more than half of the 4.4 and 4.2 points (\cref{fig:ablations}b). Without status information, the model may incorrectly treat negated findings such as ``no pleural effusion'' as positive training examples, because text embeddings struggle to distinguish a finding from its negation. This is particularly damaging for findings that radiologists frequently rule out (\cref{tab:supp-perfinding}). Further, we observe that simply excluding negated mentions performs almost as well as using them as explicit negative targets, suggesting that avoiding contradictory supervision is more important than learning from every negative statement. Crucially, the performance improvement is robust against imperfect extraction: even when a fifth of the statuses are deliberately flipped, most of the improvement over sentence-level training remains. A physician's review of the compiled findings supports the reliability of these weak labels (\cref{sec:supp-labelcheck}).

Another important takeaway is that the pretrained image encoder plays an important role (\cref{fig:ablations}c). Replacing FlexiCT-2D with natural-image DINOv3 consistently reduces performance. The benefit of typed findings over sentences is about the same with either encoder, so both the supervision type and the CT encoder contribute to the performance improvement.

\noindent\textbf{What the location adds.} In our experiments, location supervision has a different purpose. It adds little to classification performance (\cref{tab:ablation}), but changes where the model's query maps point. We separate its two components in \cref{tab:loc}. Without location supervision, the model already shows some spatial grounding from the finding-classification loss. Adding location words as training concepts substantially improves how well organ-name queries find their organs, even without the location loss. Adding the location loss further improves how accurately the model points to findings (+1.0 points, 95\% interval 0.2 to 1.7) and identifies the correct side of the body. Thus, the location words mainly teach anatomical concepts, while the location loss refines the spatial evidence for individual findings.

\begin{table}[tbp]
\caption{Localization on held-out ReXGroundingCT scans, for \method, for the same head with only the location loss switched off (No loss), and without location supervision (No loc.: no location loss, location words, or near-synonym rule). Masks set the Dice threshold and train the last row. Side: balanced accuracy. Sentences: a head trained on sentences.}
\label{tab:loc}
\centering
\footnotesize
\setlength{\tabcolsep}{3pt}
\begin{tabularx}{\columnwidth}{@{}Lcccl@{}}
\toprule
Test & \method & No loss & No loc. & Reference \\
\midrule
Best token on the mask & 37.3\% & 36.3\% & 35.6\% & random 0.3\% \\
Best token in named lobe & 61\% & 61\% & 60\% & chance 20\% \\
Best token on named side & 81\% & 79\% & 78\% & chance 50\% \\
Organ words, 24 structures & 58\% & 60\% & 42\% & sentences 53\% \\
Lobe names, among the five & 92\% & 80\% & 24\% & sentences 57\% \\
Dice, GLINT-CT's 50 cases & .137 & .135 & .126 & GLINT-CT .128 \\
U-Net, 300 masks (Dice) & .157 & .149 & .147 & map alone .139 \\
\bottomrule
\end{tabularx}
\end{table}

\begin{figure*}[t]
\centering
\includegraphics{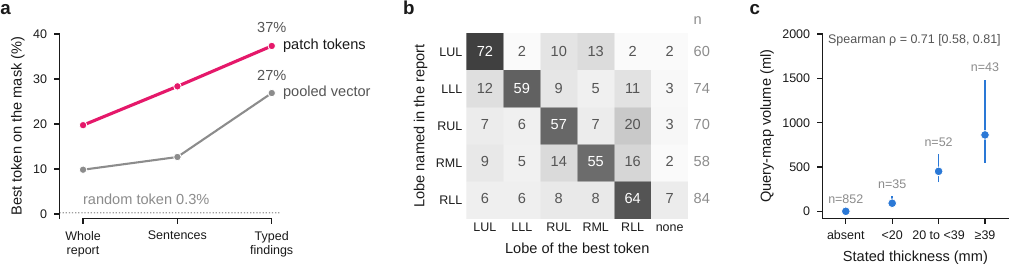}
\caption{(a) Share of 1,344 held-out ReXGroundingCT findings in 542 scans whose best-scoring token overlaps the radiologist's mask, for the six designs of \cref{fig:ablations}a (random token: 0.3\%).
(b) Lobe of the best token against the lobe the report names (row percentages, number of findings at right). LUL to RLL are the left upper and lower and the right upper, middle, and lower lobes, and ``none'' is a token outside the lungs.
(c) Volume of the pleural-effusion query map against the stated effusion thickness (median and interquartile range). Training never uses a size, and ``absent'' is left out of $\rho$.}
\label{fig:localisation}
\end{figure*}

\begin{figure*}[t]
\centering
\includegraphics{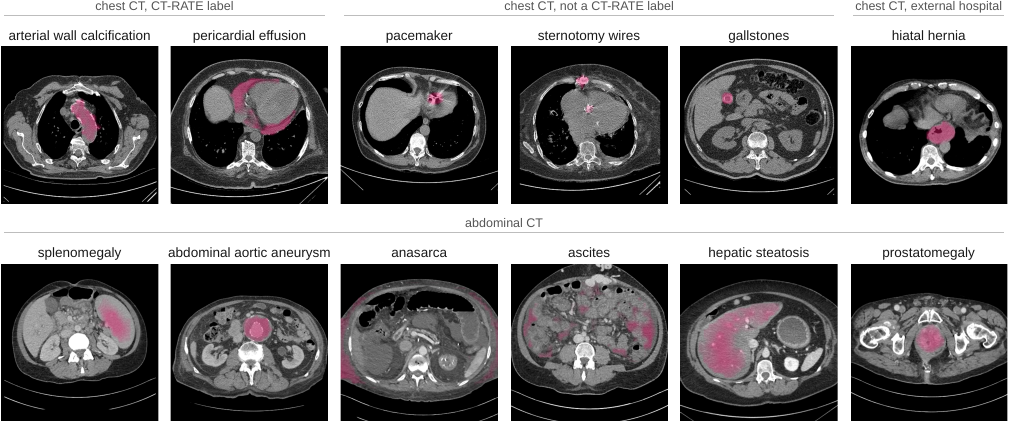}
\caption{Query maps on held-out scans. Each tile is the axial slice through the query map's highest-cosine token (magenta, fading in from cosine .40 to .65), in radiological orientation.
Top row: CT-RATE validation, queried by CT-RATE labels and other findings, and RAD-ChestCT. Bottom row: Merlin test studies.
Each tile shows the highest-scoring scan whose labels or report mark the finding present.}
\label{fig:cases}
\end{figure*}

Finally, we compare typed findings with directly supervised benchmark labels using the same head (\cref{tab:ablation}). We see that the label-trained head is slightly stronger on CT-RATE, while our \method outperforms it on RAD-ChestCT. Most of the advantage of our \method on RAD-ChestCT comes from pulmonary fibrotic sequela, which is labeled differently across the two datasets. On the other 15 findings, both models perform similarly. This suggests that findings automatically extracted from radiology reports can provide supervision comparable to predefined benchmark labels, while allowing the model to learn a much broader vocabulary of findings without specifying the target classes in advance.

\subsection{Localization from the query map}

Since our \method matches finding names to individual CT tokens, it produces a spatial query map without training on segmentation masks. We test whether these maps actually point to the findings (\cref{tab:loc,fig:localisation}), with examples in \cref{fig:cases}.

On 542 held-out ReXGroundingCT scans~\cite{rexgroundingct}, the highest-scoring token overlaps the radiologist's mask for 37\% of findings, compared with 0.3\% for a random token. Such localization performance improves alongside classification as we move from whole reports to sentences and typed findings (\cref{fig:localisation}a). For segmentation, we evaluate on the same 50 cases as GLINT-CT~\cite{glintct}, using a cosine threshold selected on 300 masked training scans. The query maps achieve a Dice score of .137, slightly above GLINT-CT's .128, but remain well below the supervised VoxTell~\cite{voxtell} (.282).

We observed that the spatial query maps also carry anatomical information. Notably, our \method locates 24 anatomical structures by name, without ever training on organ masks. Specifically, the highest-scoring token falls inside the correct organ in 58\% of scans, compared with 42\% without location supervision (\cref{fig:teaser}b). For findings whose reports specify a lung lobe, the best token falls in the correct lobe 61\% of the time, and identifies the correct side with 81\% balanced accuracy (\cref{fig:localisation}b).

Beyond location, the query maps also reflect the extent of a finding. For pleural effusion, the volume selected by a fixed cosine threshold correlates with the thickness reported by the radiologist (Spearman $\rho=.71$), even though size is never used during training (\cref{fig:localisation}c). Small nodules illustrate the opposite limitation: the model can locate them, but cannot accurately trace their boundaries. Their localization accuracy reaches 38\%, while Dice remains at .05 because the 12\,mm tokens are larger than most nodules.

The query maps can also guide a small supervised segmentation model. With 300 masked scans, a 3D U-Net~\cite{cciccek20163d} using the CT and query map improves Dice from .139 for the map alone to .157 on the 542 held-out scans (\cref{tab:loc}). Nevertheless, we find that the maps are more useful for locating findings than for outlining their boundary precisely.

\noindent\textbf{Limitations.} All comparisons hold the CT encoder frozen, and the text encoder is trained only as Qwen3-Embedding-0.6B with LoRA (\cref{sec:supp-trained}). The lead over published zero-shot models needs the CT encoder. \method scores findings by name: with the sentence-pair prompt of CT-CLIP it scores 80.3 and 74.4, and report retrieval is not tested. The weak labels come from one language model, and one reader checked 500 of their sentences, from CT-RATE only. The query map points at a finding more often than it outlines it.

\section{Conclusion}
\label{sec:conclusion}

A radiology report lists findings about one scan, each with a status and often a location; \method uses it in this form. Typed findings on the patch tokens of frozen encoders let a linear head lead zero-shot models that train their encoders, on in-domain and external chest CT, and on abdominal CT. On frozen encoders, sentences add 20 points over whole reports in-domain and typed findings 4 more, over half of it via the status. Patch tokens and the CT encoder add 2 to 6 each. Asking the stated location at a finding's own tokens moves where the model points. Typed findings make the supervision of a 3D vision model explicit and inspectable, so what a model learns from a report becomes a design choice. When a text is a list of claims, the claim is the unit to supervise.

\section*{Acknowledgments}
This work was supported through the Bavarian AI Foundation Model Initiative, funded by the Bavarian State Ministry of Science and the Arts (StMWK).
High-performance computing resources were provided by the Erlangen National High Performance Computing Center (NHR@FAU) at Friedrich-Alexander-Universität Erlangen-Nürnberg under NHR projects b180dc, y100dc.
NHR is funded by federal and Bavarian state authorities.
The NHR@FAU hardware is partly funded by the German Research Foundation (DFG; grant 440719683).
J.Wa.\ and B.K. are supported by ERC MIA-NORMAL 101083647, 101326871 and DFG 513220538, 512819079, 440719683, CL 552/1-1.

{\small
\bibliographystyle{ieeenat_fullname}
\bibliography{references}
}

\clearpage
\onecolumn
\appendix
\renewcommand{\thetable}{S\arabic{table}}
\renewcommand{\thefigure}{S\arabic{figure}}
\setcounter{table}{0}
\setcounter{figure}{0}
\begin{center}
{\Large\bfseries Supplementary material}
\end{center}

\section{The compiler}
\label{sec:supp-compiler}

The compiler is Qwen3-30B-A3B-Instruct-2507~\cite{qwen3}, run with greedy decoding.
Every unique sentence of the Findings and Impression sections of CT-RATE is compiled once (and training uses the typed findings of the training reports only).
Each call to the compiler is a chat of four turns, namely the system prompt, the worked example as a user turn, its answer as an assistant turn, and 30 sentences to compile.
If an answer is cut off at the length limit or is not valid JSON, its complete entries are kept and the remaining sentences are sent again.

\noindent\textbf{System prompt (chest CT).}
\begin{list}{}{\setlength{\leftmargin}{\dimexpr\leftmargini+1em\relax}\setlength{\rightmargin}{\leftmargini}\setlength{\itemindent}{-1em}\setlength{\labelwidth}{0pt}\setlength{\labelsep}{0pt}\setlength{\itemsep}{1pt}\setlength{\parsep}{0pt}}\small
\item You convert chest CT report sentences into structured facts. For each numbered sentence, list every finding it states.
\item Use the fields: finding (short canonical noun phrase, lowercase, singular, no negation words, e.g.\ `pleural effusion', `lung nodule', `emphysema', `normal heart size'), present (`yes', `no' or `uncertain'), location (anatomic location or null), detail (size, number, density, chronicity or other qualifiers, or null).
\item A sentence that negates a finding gives present `no' (e.g.\ `No pleural effusion' $\rightarrow$ finding `pleural effusion', present `no').
\item A sentence stating that a structure is normal gives a finding naming the normal state (e.g.\ `heart size is normal' $\rightarrow$ finding `normal heart size', present `yes').
\item Split coordinated findings into separate facts, also when they are joined by hyphens or slashes (e.g.\ `no mass nodule-infiltration' gives three facts), and keep every clause of a sentence (e.g.\ `X was normal and no Y was detected' gives two facts).
\item Never give a bare structure name: when a sentence states a structure's state only with a verb or adjective, name the state in the finding, with `normal' for a normal state also when the report says `natural', `open' or `preserved' (e.g.\ `heart size increased' $\rightarrow$ `increased heart size'; `trachea and both main bronchi are open' $\rightarrow$ `normal trachea', `normal main bronchus').
\item CTO means cardiothoracic ratio.
\item A heading or viewing condition that names no finding (e.g.\ `when examined in the lung parenchyma window;', `as far as can be seen;') gives an empty list; every other sentence lists the findings it mentions.
\item Answer with one JSON object on a single line mapping each sentence number (as a string) to a list of facts, and nothing else.
\end{list}

\noindent\textbf{Worked example (chest CT).} \begin{quote}\small
\textbf{1.} Pleural effusion-thickening was not detected bilaterally.\\
\textbf{2.} A 6 mm nodule is observed in the superior segment of the right lower lobe.\\
\textbf{3.} Heart contour and size are normal.\\
\textbf{4.} When examined in the mediastinal window;\\
\textbf{5.} No mass nodule-infiltration was detected in both lungs.\\
\textbf{6.} Thoracic esophagus calibration was normal and no wall thickening was detected.\\
\textbf{7.} Heart size increased, CTO is increased.\\
\textbf{8.} Trachea and both main bronchi are open.\\
\textbf{9.} Calibration of thoracic main vascular structures is natural.
\end{quote}
\begin{list}{}{\setlength{\leftmargin}{\leftmargini}\setlength{\rightmargin}{0pt}}\item[]
\sffamily\fontsize{7.5}{9.5}\selectfont\frenchspacing
\begin{tabular}{@{}l@{}l@{\ }l@{\ }l@{\ }l@{}}
\wxp{\{}\wxn{1} & \wxf{pleural effusion}{no}{bilateral}{\wxnull\wxp{\},}} \\
 & \wxf{pleural thickening}{no}{bilateral}{\wxnull\wxp{\}],}} \\[1.5pt]
\phantom{\{}\wxn{2} & \wxf{lung nodule}{yes}{right lower lobe superior segment}{\wxs{wxink}{6 mm}\wxp{\}],}} \\[1.5pt]
\phantom{\{}\wxn{3} & \wxf{normal heart contour}{yes}{heart}{\wxnull\wxp{\},}} \\
 & \wxf{normal heart size}{yes}{heart}{\wxnull\wxp{\}],}} \\[1.5pt]
\multicolumn{5}{@{}l@{}}{\phantom{\{}\wxn{4}\wxp{],}} \\[1.5pt]
\phantom{\{}\wxn{5} & \wxf{mass}{no}{both lungs}{\wxnull\wxp{\},}} \\
 & \wxf{lung nodule}{no}{both lungs}{\wxnull\wxp{\},}} \\
 & \wxf{infiltration}{no}{both lungs}{\wxnull\wxp{\}],}} \\[1.5pt]
\phantom{\{}\wxn{6} & \wxf{normal thoracic esophagus calibration}{yes}{thoracic esophagus}{\wxnull\wxp{\},}} \\
 & \wxf{esophageal wall thickening}{no}{thoracic esophagus}{\wxnull\wxp{\}],}} \\[1.5pt]
\phantom{\{}\wxn{7} & \wxf{increased heart size}{yes}{heart}{\wxnull\wxp{\},}} \\
 & \wxf{increased cardiothoracic ratio}{yes}{heart}{\wxnull\wxp{\}],}} \\[1.5pt]
\phantom{\{}\wxn{8} & \wxf{normal trachea}{yes}{trachea}{\wxnull\wxp{\},}} \\
 & \wxf{normal main bronchus}{yes}{both main bronchi}{\wxnull\wxp{\}],}} \\[1.5pt]
\phantom{\{}\wxn{9} & \wxf{normal vascular calibration}{yes}{thoracic main vascular structures}{\wxnull\wxp{\}]\}}}
\end{tabular}
\end{list}

\noindent\textbf{Abdominal CT (Merlin).} The system prompt says ``abdominal CT report sentences'' where the chest prompt says ``chest CT report sentences'', and any other rule is unchanged.

\begin{samepage}
\noindent\textbf{Worked example (abdominal CT).} \begin{quote}\small
\textbf{1.} kidneys and ureters: no hydronephrosis/hydroureter bilaterally.\\
\textbf{2.} a 1.2 cm hypodense lesion is seen in hepatic segment 7.\\
\textbf{3.} spleen: normal in size and contour.\\
\textbf{4.} impression: 1.\\
\textbf{5.} peritoneum: no ascites/free air/fluid collection.\\
\textbf{6.} the gallbladder is normal and there is no biliary ductal dilatation.\\
\textbf{7.} liver size is increased, measuring 21 cm.\\
\textbf{8.} the portal vein and hepatic veins are patent.\\
\textbf{9.} the caliber of the abdominal aorta and iliac arteries is preserved.
\end{quote}
\end{samepage}
\begin{list}{}{\setlength{\leftmargin}{\leftmargini}\setlength{\rightmargin}{0pt}}\item[]
\sffamily\fontsize{7.5}{9.5}\selectfont\frenchspacing
\begin{tabular}{@{}l@{}l@{\ }l@{\ }l@{\ }l@{}}
\wxp{\{}\wxn{1} & \wxf{hydronephrosis}{no}{bilateral}{\wxnull\wxp{\},}} \\
 & \wxf{hydroureter}{no}{bilateral}{\wxnull\wxp{\}],}} \\[1.5pt]
\phantom{\{}\wxn{2} & \wxf{liver lesion}{yes}{hepatic segment 7}{\wxs{wxink}{1.2 cm, hypodense}\wxp{\}],}} \\[1.5pt]
\phantom{\{}\wxn{3} & \wxf{normal spleen size}{yes}{spleen}{\wxnull\wxp{\},}} \\
 & \wxf{normal spleen contour}{yes}{spleen}{\wxnull\wxp{\}],}} \\[1.5pt]
\multicolumn{5}{@{}l@{}}{\phantom{\{}\wxn{4}\wxp{],}} \\[1.5pt]
\phantom{\{}\wxn{5} & \wxf{ascites}{no}{peritoneum}{\wxnull\wxp{\},}} \\
 & \wxf{free air}{no}{peritoneum}{\wxnull\wxp{\},}} \\
 & \wxf{fluid collection}{no}{peritoneum}{\wxnull\wxp{\}],}} \\[1.5pt]
\phantom{\{}\wxn{6} & \wxf{normal gallbladder}{yes}{gallbladder}{\wxnull\wxp{\},}} \\
 & \wxf{biliary ductal dilatation}{no}{biliary tree}{\wxnull\wxp{\}],}} \\[1.5pt]
\phantom{\{}\wxn{7} & \wxf{increased liver size}{yes}{liver}{\wxs{wxink}{21 cm}\wxp{\}],}} \\[1.5pt]
\phantom{\{}\wxn{8} & \wxf{normal portal vein}{yes}{portal vein}{\wxnull\wxp{\},}} \\
 & \wxf{normal hepatic vein}{yes}{hepatic veins}{\wxnull\wxp{\}],}} \\[1.5pt]
\phantom{\{}\wxn{9} & \wxf{normal vascular caliber}{yes}{abdominal aorta and iliac arteries}{\wxnull\wxp{\}]\}}}
\end{tabular}
\end{list}

\noindent\textbf{From typed findings to training targets.} The finding name and the status (the field \texttt{present}) enter training through the scan-level status of \cref{sec:method}, and the location through the location words and the location loss (\cref{sec:supp-location}). The detail is kept for the size test of \cref{sec:experiments}.

\section{The concept inventory}
\label{sec:supp-concepts}

\begin{figure}[h]
\centering
\includegraphics{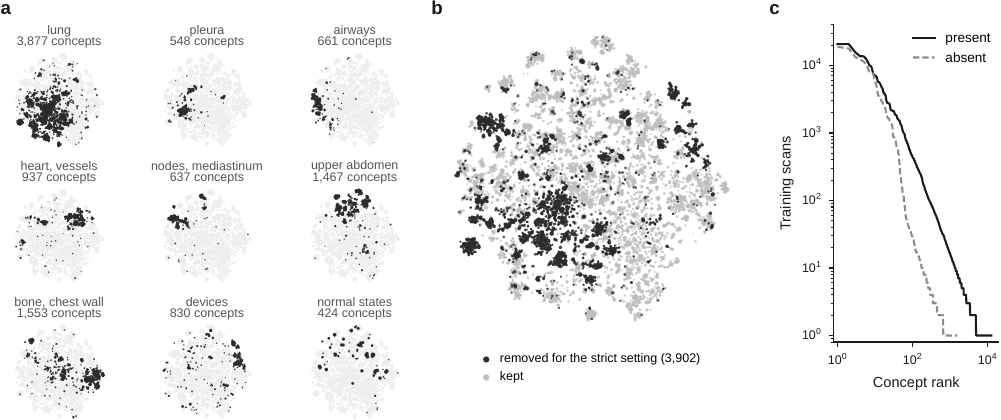}
\caption{Concepts compiled from the CT-RATE reports.
(a) t-SNE~\cite{van2008visualizing} of the 14,619 concept names in the text embedding space, one map per body region that a keyword list assigns to the names. The concepts of the region are black, and point area grows with the number of training scans in which a concept is present or absent.
(b) The same map, with the 3,902 concepts removed for the strict setting (\cref{sec:supp-strict}) in black.
(c) Training scans per concept for present and absent targets, with concepts ranked by each.}
\label{fig:concepts}
\end{figure}

The compiler returned 20,577 distinct finding names on the CT-RATE reports, and the merge of \cref{sec:method} turns them into 14,619 concepts, 14,630 with the location words of \cref{sec:supp-location}.
A run without any merging has 20,588 concepts and similar scores (\cref{tab:supp-location}).
Across the training scans, the 14,619 concepts hold 363,789 present and 175,915 absent targets.
Of these concepts, 12,702 are present in at least one training scan and 880 in at least 20.
\Cref{tab:supp-top} lists the most frequent concepts, and \cref{fig:concepts} maps all of them.
The vocabulary spans the anatomy of a chest CT, devices, and normal states, with a long tail of rare concepts (\cref{fig:concepts}a,c).
Normal states (``normal trachea'', ``normal heart size'') dominate the present targets, because reports state them in most scans. Findings that radiologists rule out by habit dominate the absent targets.

\begin{table}[h]
\caption{Most frequent concepts in the CT-RATE training reports, as present and as absent targets (number of training scans).}
\label{tab:supp-top}
\centering
\small
\begin{tabular}{@{}lrr@{\hspace{2.5em}}lrr@{}}
\toprule
\multicolumn{3}{c}{Most often present} & \multicolumn{3}{c}{Most often absent} \\
Concept & present & absent & Concept & present & absent \\
\midrule
normal trachea & 20,687 & 0 & esophageal wall thickening & 61 & 18,832 \\
normal main bronchus & 20,645 & 0 & pericardial effusion & 1,553 & 17,781 \\
normal heart contour & 15,632 & 0 & pleural effusion & 2,797 & 13,314 \\
normal thoracic esophagus calibration & 13,660 & 0 & pericardial thickening & 77 & 12,276 \\
normal heart size & 13,648 & 8 & enlarged lymph node & 152 & 11,394 \\
normal mediastinal main vascular structures & 12,173 & 0 & space occupying lesion & 71 & 10,204 \\
normal adrenal gland & 10,147 & 0 & lytic destructive lesion & 9 & 8,120 \\
normal vertebral corpus height & 9,639 & 0 & pleural thickening & 620 & 8,036 \\
normal upper abdominal organ appearance & 7,582 & 0 & occlusive pathology & 1 & 7,484 \\
lung nodule & 7,043 & 3,376 & mass & 315 & 5,807 \\
normal bone structure & 6,843 & 0 & infiltrative lesion & 9 & 4,989 \\
normal vascular calibration & 5,689 & 0 & mass lesion & 270 & 3,681 \\
\bottomrule
\end{tabular}
\end{table}

\section{Target findings removed from training}
\label{sec:supp-strict}

\noindent\textbf{What is removed.} The compiled vocabulary covers the target CT-RATE findings: 14 of the 18 label names are names of concepts with at least 770 positive training scans. Thus, the strict setting removes every concept whose name matches the pattern of a target finding.
The patterns cover the name of the finding, its synonyms, subtypes, and qualified forms. For example, the pattern of lung opacity also matches ground-glass opacity, infiltrate, crazy paving, and dependent density.
Some patterns leave out the organ, which is stricter than the finding itself: the pattern ``nodul'' also removes thyroid nodules, for example.
Synonyms that the patterns missed were found by a screen of the concept embeddings.
The rule removes 3,902 of the 14,619 finding concepts, which hold 30.3\% of the present targets.

The removed concepts form clusters across the whole map, and related concepts that stay in training lie next to them (\cref{fig:concepts}b).
A concept name can be generic (``fluid collection'' with the pleural space as its location), so a second filter also drops every remaining typed finding whose name, location, and detail together match a target pattern. This removes another 1.7\% of the present targets, 32\% in all.
A removed concept enters training neither as present nor  absent or unmentioned. The scans stay in training with their other concepts.

\noindent\textbf{Example of pleural effusion.} The pattern removes 69 concepts, among them pleural effusion (present in 2,797 training scans), bilateral pleural effusion (132), minimal pleural effusion (91), loculated pleural effusion (34), and pleural fluid (26).
Among concepts present in at least 20 training scans, the closest that stay are abdominal effusion (text cosine .83), fluid collection (.81), perihepatic effusion (.81), pulmonary edema (.81), and pneumothorax (.80).
The second filter also drops the mentions of these concepts that name the target, which are 4 of the 30 present mentions of fluid collection and 12 of the 93 of pulmonary edema, such as ``compatible with pleural effusion''.

\noindent\textbf{Random controls.} Removing 32\% of the present targets could cost accuracy by itself.
Two controls remove a similar share of the training text, chosen at random instead of by target.
The first removes present and absent mentions, in the shares that the removed concepts hold (30.3\% of the present targets), so every concept stays in training.
The second removes whole concepts, 3,902 that name no target and match the removed concepts in training frequency.
Each control costs at most 0.2 points, while the strict setting costs 9.4 to 10.4 (\cref{tab:supp-strict}).

\noindent\textbf{Findings held out by CT-DiagVLM.} CT-DiagVLM~\cite{ctdiagvlm} holds six of its 16 CT-RATE findings out of training, which its Appendix A.4 gives only by index: five lung findings, the last five of CT-RATE's label order (mosaic attenuation pattern, peribronchial thickening, consolidation, bronchiectasis, and interlobular septal thickening), and one heart finding. It scores 83.1 on these six. With each of the three heart findings in turn as the sixth (cardiomegaly, pericardial effusion, and coronary artery wall calcification), \method trained in the strict setting scores 76.0 to 77.5. Note that CT-DiagVLM keeps its other ten findings in training; our strict setting removes all 18.

\begin{table}[h]
\caption{Strict setting and random controls (zero-shot macro AUROC, label-name queries). \method is mean of three seeds, other rows are one run. CT-RATE 16 and RAD 14 are the 16 and 14 findings that fVLM reports. fVLM needs organ masks at test time.}
\label{tab:supp-strict}
\centering
\small
\begin{tabular}{@{}lcccc@{}}
\toprule
Training text & CT-RATE 18 & CT-RATE 16 & RAD 16 & RAD 14 \\
\midrule
All concepts (\method) & 87.7 & 88.3 & 82.8 & 83.3 \\
Random control, same share of mentions removed & 87.5 & 88.1 & 82.6 & 83.2 \\
Random control, 3,902 frequency-matched other concepts removed & 87.7 & 88.2 & 82.9 & 83.3 \\
Strict, every target concept removed & 77.9 & 77.9 & 73.4 & 73.9 \\
\midrule
fVLM~\cite{fvlm} & - & 77.8 & - & 68.0 \\
\bottomrule
\end{tabular}
\end{table}

\section{Results per finding}
\label{sec:supp-perfinding}

\begin{figure}[tp]
\centering
\includegraphics[trim=27.5pt 0pt 0pt 0pt]{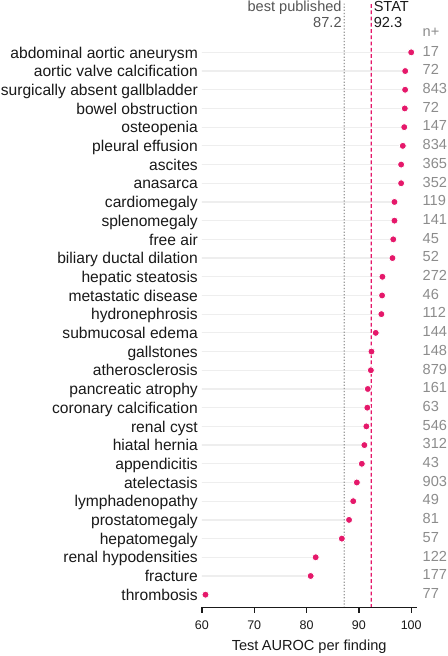}%
\caption{The chest recipe on abdominal CT (Merlin).
Test AUROC per finding on 5,125 studies, with each finding scored on the studies whose report mentions it. n+ is the number of positive studies.
Dashed lines mark the macro AUROC of \method and of the prior state-of-the-art zero-shot model, FlexiCT-3D-VLM, which also trains on Merlin's reports.}
\label{fig:merlin}
\end{figure}

\begin{figure}[tp]
\centering
\includegraphics{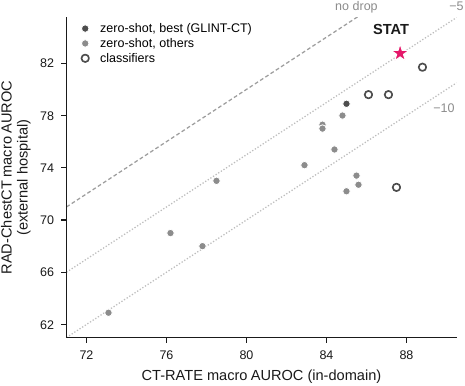}
\caption{Drop between hospitals.
In-domain (CT-RATE) against external-hospital (RAD-ChestCT) macro AUROC for every zero-shot model with both numbers (filled) and for classifiers (open), including the same head as \method on CT-RATE's labels (88.8, 81.7).
Dashed line: no drop. Dotted lines: drops of 5 and 10 points.}
\label{fig:supp-transfer}
\end{figure}

\Cref{tab:supp-perfinding} gives the AUROC of each finding for the main design steps on chest CT, \cref{fig:merlin} gives the AUROC of each finding on Merlin, and \cref{fig:supp-transfer} compares in-domain with external-hospital scores.
Two of the 18 findings stay at 90 or above in the strict setting, because related concepts remain in training. For 14 of the 18 targets, the nearest kept concept has a text cosine of at least .80.

\begin{table}[tp]
\caption{AUROC per finding. Sent.: sentences. No st.: names without status. Strict: target findings removed from training. Labels: the same head on CT-RATE's 18 labels. \method and Labels are means of three seeds. Each column uses the query form of its row in \cref{tab:ablation}. ``-'' marks a finding that RAD-ChestCT does not label.}
\label{tab:supp-perfinding}
\centering
\small
\setlength{\tabcolsep}{4pt}
\input{tab_perfinding}
\end{table}

\section{Compiled targets against CT-RATE's labels}
\label{sec:supp-agreement}

CT-RATE's labels come from the same reports through a text classifier~\cite{ctclip}, so agreement with them checks the compiler rather than measuring its accuracy.
\Cref{tab:supp-agreement} gives Cohen's kappa~\cite{cohen1960coefficient} on the 24,128 scans of CT-RATE's training split, each with one report and about two reconstructed volumes, in two readings, by concept name alone and by the whole typed finding with its location and detail. The head trains on the 45,850 volumes of the 23,469 scans that remain after the hold-out. The two calcification labels score low by name alone, because the compiler writes ``atheroma plaque'' and puts the vessel in the location field. Read with location and detail, they reach .79 and .84. For cardiomegaly, the name pattern covers both word orders (``increased heart size'').
Lymphadenopathy stays low because CT-RATE's label is positive in 25\% of scans and the compiled concept in 6\%. Medical material falls to .59 with location and detail, where its pattern also matches unrelated text.

{\setlength{\intextsep}{9pt}%
\begin{table}[h]
\caption{Compiled targets against CT-RATE's labels (Cohen's kappa).}
\label{tab:supp-agreement}
\centering
\small
\input{tab_agreement}
\end{table}}

\section{Label check}
\label{sec:supp-labelcheck}

One of the authors, a physician, read 500 sentences drawn at random from the Findings and Impression sections of CT-RATE training reports.
Frequent sentences were drawn several times, so the 500 draws contain 343 different sentences.
Each percentage in \cref{tab:supp-labelcheck} counts a typed finding once per draw of its sentence, as often as training sees it.
Counted this way, the 500 draws hold 709 typed findings.
The reader saw each sentence with its report, and every typed finding with its name, merged concept, status, location, and detail.
A typed finding is judged as training reads it: the name through its concept, and the status.
The detail is never read, and the location only as a lung lobe, a side, or an organ of a present finding.
A finding is accepted as an image finding if it names something the scan shows: an abnormality, the normal state of a structure, or a post-treatment state.
Statements about the examination, recommendations, history, and comparisons are not considered image findings.
``Consistent with'', ``compatible with'', and ``in favor of'' count as present; ``may'', ``possible'', ``?'', and differential diagnoses count as uncertain.
A present finding whose qualifier sits in the detail (``calcified'') is correct but less specific.
An absent finding is wrong when its bare name is a structure the scan normally shows, e.g.\ ``no lymph node of pathological size'' typed as an absent lymph node; half of the wrong targets are of this kind.

{\setlength{\intextsep}{9pt}%
\begin{table}[h]
\caption{Label check: 500 random CT-RATE training sentences read by a physician.}
\label{tab:supp-labelcheck}
\centering
\small
\begin{tabular*}{0.62\textwidth}{@{\extracolsep{\fill}}lr@{}}\toprule
Typed findings (709) & \% [95\% CI] \\\midrule
correct & 92.2 [90.0, 94.0] \\
correct, less specific & 0.8 [0.4, 1.8] \\
not an image finding & 2.8 [1.8, 4.3] \\
wrong target & 4.1 [2.9, 5.8] \\
\quad lymph node typed without ``pathological'' & 2.1 [1.3, 3.5] \\
\quad hedge or history typed as present & 0.8 [0.4, 1.8] \\
\quad no finding named, or not stated & 0.7 [0.3, 1.6] \\
\quad merged with a different finding & 0.4 [0.1, 1.2] \\\midrule
stated findings typed (recall) & 97.9 [96.6, 98.7] \\
correct location, of 696 with a location & 97.4 [95.9, 98.4] \\\bottomrule
\end{tabular*}
\end{table}}

\section{Training and evaluation details}
\label{sec:supp-details}

\noindent\textbf{Data.} Each axial slice is encoded at 0.75\,mm in-plane and 1.5\,mm between slices. A scan keeps at most 360\,mm along the body axis, centered on the slices with tissue, and 600\,mm on Merlin. The released Merlin findings text includes the impression in 79\% of reports, so \method trains on findings and impression where present.

\noindent\textbf{Training.} The head trains with AdamW~\cite{loshchilov2017decoupled} (learning rate $10^{-3}$, weight decay $10^{-4}$, batch size 8 scans), 500 warm-up steps, and a cosine decay to 5\% of the peak.
The location loss uses the $K=128$ best tokens of a finding and the weight $\lambda=.3$.
The temperature $\tau$ starts at .05, the scale $t$ at 10, and the bias $b$ at $-10$.
Log-sum-exp pooling runs over all tokens of a scan, which are the patch tokens and one CLS token per slice.
The pooled head replaces it by attention pooling with one learned query~\cite{ilse2018attention}, which gives one vector per scan. Its query map is still the cosine of each projected token with the name's embedding $q_n$.

\noindent\textbf{Cells of the grid in \cref{fig:ablations}a.} In the whole-report cells, each report is one text. The positive of a scan is its own report, and the negatives are the reports of the other scans in the batch and 256 drawn training reports. In the sentence cells, sentences with a text cosine of at least .95 form one group, and the group is the unit. The positives of a scan are the groups of its report, and the negatives are the other groups in the batch and 256 groups sampled by frequency. ``Sentences with status'' keeps only the sentences that state a present finding as targets. It scores 83.6 on CT-RATE and 78.5 on RAD-ChestCT, against 83.3 and 78.6 for sentences without status (\cref{fig:ablations}b). ``Finding names without status'' counts every mention as present, so that ``no X'' trains X as present, and uncertain mentions stay ignored. In \cref{tab:ablation}, the row ``no X'' left out of training removes the absent mentions, so that the finding is unmentioned in that scan and trains as a weak negative. The row ``Linear probe on the pooled vector'' fits one logistic regression per label on the mean of a scan's patch tokens.

\noindent\textbf{Evaluation.} \method is queried by label name. With the pair prompt of CT-CLIP (``X is present.'' against ``X is not present.''), it scores 80.3 on CT-RATE and 74.4 on RAD-ChestCT, because the head never trained on negated text.
Jolia~\cite{jolia} scores Merlin with eight fixed prompt pairs per finding. Under this protocol Jolia scores 78.0, the Merlin model 77.7, and \method 89.2. For the 43 findings of the three benchmarks, ``no X'' and ``X'' have a median cosine of .84 in \method's frozen text encoder $F$, against .55 between different findings, and ``no X'' is nearest to ``X'' itself for all 43.
The three query forms of \cref{tab:ablation} are ``X is present.'', ``X is present.'' minus ``X is not present.'', and the label name ``X''. Models trained on sentences do best with ``X is present.'', and \method with the plain label name as trained.

RAD-ChestCT uses the 16 findings of CT-CLIP. Arterial wall calcification is scored per scan with the higher of the arterial and coronary queries, because the dataset has one calcification label.

Each ReXGroundingCT finding is a report sentence with a mask, and \method queries it by the finding names that the compiler extracts from that sentence, such as ``air cyst'' from ``Air cyst at the anterobasal level of the right lower lobe''. The query map and the best token use the patch tokens only. The 542 held-out scans are the 200 validation scans of ReXGroundingCT, which come from CT-RATE's training split, and 342 CT-RATE validation scans. The CT encoder saw the images of the 200 in pretraining, without masks or reports. On the 342 scans that the encoder never saw, the best token lies on the mask for 35\% of findings over the three seeds of \method, and the thresholded query map reaches a Dice score of .140, against 44\% and .137 on the 200 that it saw. On the 300 masked scans that set the Dice threshold, whose images the encoder saw and whose reports trained the head, the best token lies on the mask for 25\% of findings, and the thresholded query map reaches a Dice score of .096.   
Lobes, lung sides, and organ masks come from CADS segmentations~\cite{cads}.
The organ words are 24 location words with a CADS mask: the five lobes and 19 organs. Each is queried by its plain name, on the held-out scans in which CADS segments it (209 to 542 scans per word). Chance is the share of a scan's patch tokens that lie on the mask, averaged over the same scans.
The effusion query-map volume counts tokens with a cosine of at least .55.
Nodule scores on LUNA16 and LUNA25 use log-sum-exp pooling over the 3$\times$3 tokens around the annotated nodule center.
On LIDC~\cite{lidc}, the query ``lung nodule'' is scored against the radiologists' nodule outlines by the share of scans whose best token lies on a nodule and by the token AUROC, which separates the nodule tokens of a scan from its other tokens.
The U-Net of \cref{tab:loc} takes the CT and the query map as input. With the text embedding of the finding in place of the query map, it reaches a Dice score of only .068 on the held-out scans. On GLINT-CT's 50 cases, the U-Net reaches .127, below the query map alone (\cref{tab:loc}).

\section{Training the text encoder}
\label{sec:supp-trained}

The grid in \cref{fig:ablations}a holds both encoders frozen. \Cref{tab:supp-trained} repeats the step from sentences to typed findings with a trained text encoder.
The text encoder is Qwen3-Embedding-0.6B, because the 8B model is too slow to train at every step.
LoRA~\cite{lora} of rank 16 on the query and value projections of its 28 blocks (2.3M parameters) trains together with the head from step 2,000, at a learning rate of $2\times10^{-4}$.
The head, the CT tokens, and the 30k steps are those of \cref{tab:ablation}.
A sentence group is represented by its most frequent sentence, because the mean embedding of all its sentences, which the grid in \cref{fig:ablations}a uses, is too slow to recompute at every step.
With the frozen encoder, the most frequent sentence scores at least as high as the mean on the first seed (81.2 vs.\ 81.1 on CT-RATE, 76.3 vs.\ 75.6 on RAD-ChestCT).

The trained sentence runs overfit: their validation AUROC peaks between steps 6k and 16k and lies below that of the frozen runs at step 30k.
We therefore read every run at the step with its highest validation AUROC.
The validation patients belong to CT-RATE's test set, so this reading favors a run on CT-RATE.
Typed findings then outperform sentences by 5.0 and 5.3 points with the trained encoder, against 5.4 and 5.7 with the frozen one.
Training the text encoder raises sentences by 0.6 and 0.5 points and typed findings by at most 0.2.
At step 30k, as in \cref{tab:ablation}, the lead is 6.6 and 7.5 points with the trained encoder, against 5.7 and 5.8 with the frozen one.

The trained encoder does not set a negated finding apart either.
At the best validation step, the encoder trained on sentences lowers the median cosine of ``no X'' and ``X'' over the 43 benchmark findings from .90 to .79, but that of two different findings from .62 to .34, and ``no X'' stays nearest to ``X'' itself for 42 to 43 of the 43.
At step 30k, the two cosines are .65 and .24, and ``no X'' stays nearest for 41 to 42.

\begin{table}[h]
\caption{Training the text encoder (macro AUROC, mean of three seeds).
Qwen3-Embedding-0.6B is frozen, or trained together with the head by LoRA~\cite{lora}.
Each run is read at the step with its highest validation AUROC, and at step 30k.
Each row reports its best of three query forms on each test set, as in \cref{tab:ablation}.}
\label{tab:supp-trained}
\centering
\small
\begin{tabular}{@{}llcccc@{}}
\toprule
 & & \multicolumn{2}{c}{Best validation step} & \multicolumn{2}{c}{Step 30k} \\
Text unit & Text encoder & CT-RATE & RAD 16 & CT-RATE & RAD 16 \\
\midrule
Sentences & frozen & 81.6 & 76.5 & 81.2 & 76.3 \\
Sentences & LoRA & 82.2 & 77.0 & 80.4 & 75.0 \\
Typed findings & frozen & 87.0 & 82.3 & 86.9 & 82.2 \\
Typed findings & LoRA & 87.2 & 82.3 & 87.0 & 82.5 \\
\midrule
\multicolumn{2}{@{}l}{Typed findings minus sentences, frozen} & $+5.4$ & $+5.7$ & $+5.7$ & $+5.8$ \\
\multicolumn{2}{@{}l}{Typed findings minus sentences, LoRA} & $+5.0$ & $+5.3$ & $+6.6$ & $+7.5$ \\
\bottomrule
\end{tabular}
\end{table}

\section{Location supervision}
\label{sec:supp-location}

\noindent\textbf{Location items.} The location field of a typed finding is free text. Patterns read from it the lung lobes (``right upper lobe''), the side (``left'', ``right'', ``bilateral''), and 25 organs. On CT-RATE's training split this gives 33,071 lobe items and 40,092 side items for present findings other than normal states.

\noindent\textbf{Location words.} The five lobe names, ``left side'', ``right side'', and the 25 organ names are concepts of the vocabulary, 11 of them new. A location word is present in a scan when a present finding other than a normal state is located there.

\noindent\textbf{Near-synonyms.} For each concept, the near-synonym rule holds the concepts whose names have a text cosine of at least .80 with it, unless one carries a word of increase or of decrease that the other lacks. Normal states have none. 13,708 of the 14,630 concepts have at least one. Names with the word ``mass'' are kept apart from names without it, because the text encoder places ``lung mass'' next to ``lung nodule'' (cosine .88), although the two are different findings. This rule was added with the nodule malignancy scores of \cref{tab:other} in view, so these two scores are not clean tests of it.

\noindent\textbf{Effect.} The location loss at the finding's own tokens moves the finding's map, and the location words teach the organ words (\cref{tab:supp-location}). To separate the two, we switch off the location loss and keep the location words and the near-synonym rule. The best token then lies on the finding's mask less often than for \method, while the organ words keep their gain over ``name and status only''. Queried as plain words, the five lobe names find their lobe more often without the location loss than with it. This gap lies in the read-out: the location loss trains each lobe name against the other four, a question that the best token of the plain word does not ask. For the row ``Lobe names, among the five'' of \cref{tab:loc}, we ask this question as in \cref{eq:loc}, with the training setting $K=128$: among the $K$ best tokens of a lobe name, we take the token with the highest $p_i(l)$. Read this way, the lobe names find their lobe correctly more often with the location loss than without it, and in both cases far more often than for ``name and status only''. This read-out also needs no training and no mask. With $K=16$, the lobe names instead find their lobe correctly more often without the location loss than with it (80.7\% vs.\ 79.1\%).

\noindent\textbf{A softer softmax.} Queried as plain words, the lobe names find their lobe less often with the location loss than without it, because the softmax of \cref{eq:loc} runs at the sharp pooling temperature $\tau$, where training avoids confident wrong lobes by making the five lobe names nearly alike. At a temperature of 0.1, with $\lambda$ raised to 1.2 to keep the loss's pull, the lobe names stay apart: as plain words they find their lobe in 82.9\% of scans instead of 41.8\%, every best-token column of \cref{tab:supp-location} rises (fourth row), and the CT-RATE and RAD-ChestCT AUROCs stay within 0.2. All other results keep the setting of \cref{sec:experiments}.

\begin{table}[h]
\caption{Location supervision on chest CT. Upper block: means of three seeds. Lower block: the first seed of \method, and a single run on it without the name merge. Mask, Lobe, and Side: best token on the finding's mask, in the named lung lobe, and on the named side, as in \cref{tab:loc}. Organ words: the 24 structure words, each queried by its plain name. Lobe words: the five lobe names, which are part of the 24. LIDC: the query is ``lung nodule'', and the mask is the radiologists' nodule outlines.}
\label{tab:supp-location}
\centering
\small
\setlength{\tabcolsep}{5pt}
\begin{tabular*}{\textwidth}{@{\extracolsep{\fill}}lccccccccc@{}}
\toprule
 & \multicolumn{2}{c}{AUROC} & \multicolumn{5}{c}{Held-out ReXGroundingCT scans} & \multicolumn{2}{c}{LIDC nodules} \\
 & CT-RATE & RAD 16 & Mask & Lobe & Side & Organ words & Lobe words & Mask & Token AUROC \\
\midrule
Name and status only & 87.4 & 82.5 & 35.6\% & 60.1\% & 77.6\% & 42.3\% & 19.6\% & 57.7\% & 96.9 \\
\method & 87.7 & 82.8 & 37.3\% & 61.2\% & 80.6\% & 58.2\% & 41.8\% & 61.7\% & 97.6 \\
 \quad location loss off & 87.7 & 82.7 & 36.3\% & 60.7\% & 79.1\% & 60.0\% & 53.7\% & 60.0\% & 97.7 \\
 \quad softmax at 0.1, $\lambda=1.2$ & 87.7 & 82.6 & 38.2\% & 61.9\% & 82.2\% & 68.6\% & 82.9\% & 64.7\% & 97.2 \\
\midrule
\method, first seed & 87.6 & 82.8 & 37.3\% & 61.6\% & 79.8\% & 58.1\% & 41.3\% & 61.7\% & 97.4 \\
\quad names not merged & 87.6 & 82.9 & 37.1\% & 62.7\% & 79.8\% & 59.3\% & 42.7\% & 60.9\% & 97.6 \\
\bottomrule
\end{tabular*}
\end{table}

\section{Other datasets}
\label{sec:supp-other}

\Cref{tab:other} gives the zero-shot results on the four public datasets of \cref{sec:experiments}. We chose each query from the target disease of its task and fixed it before scoring: ``covid 19 pneumonia'' for COVID-19 and its severe outcome, and ``lung mass'' for nodule malignancy. Nodule scores use the annotated location.
Supervised models remain stronger on radiologist-rated nodule malignancy, with 94.4 against 92.4 on LUNA16 nodules from LIDC~\cite{lidc,fmcib} and 82 against 74 on LUNA25~\cite{luna25}.
On COVID-19 in France (STOIC~\cite{stoic}), \method scores 84.4 on the 2,000 public scans whose images our encoder saw. The challenge winners, who trained on 9,724 STOIC scans, score 84.5 to 84.9 on the challenge's hidden test set.
COVID-19 severity, an outcome that no report finding names, gives 41.3, and 16 labeled cases per class lift it to 68.8 (\cref{sec:supp-fewshot}).

\begin{table}[h]
\caption{Zero-shot AUROC on other public datasets.
Saw images: the frozen encoder saw these scans in self-supervised pretraining, without labels or reports.
LUNA16 uses the split of~\cite{fmcib}. \method is the mean of three seeds.}
\label{tab:other}
\centering
\footnotesize
\setlength{\tabcolsep}{3pt}
\begin{tabular*}{0.75\textwidth}{@{\extracolsep{\fill}}lclcl@{}}
\toprule
Dataset & Saw images & Task & \method & Best published \\
\midrule
MosMed & no & COVID-19 & \textbf{95.0} & 93.3~\cite{jin2020development}, not trained on MosMed \\
STOIC & yes & COVID-19 & 84.4 & 84.5 to 84.9~\cite{stoic}, supervised \\
STOIC & yes & severe outcome & 41.3 & 81.5~\cite{stoic}, supervised \\
LUNA16 & partly & malignancy & 92.4 & 94.4~\cite{fmcib}, fine-tuned \\
LUNA25 & likely & malignancy & 74.0 & 82~\cite{tushar2024reproducible}, not trained on LUNA25 \\
\bottomrule
\end{tabular*}
\end{table}

\section{Few-shot query choice}
\label{sec:supp-fewshot}

The queries of \cref{tab:other} were fixed before scoring.
Here the query is instead chosen with $k$ labeled cases per class, from a list of 12 to 15 candidates per task that we wrote by hand, such as ``viral pneumonia'' for COVID-19 and ``spiculated nodule'' for malignancy.
One draw takes $k$ positive and $k$ negative cases, at most one per patient, and keeps the candidate with the highest AUROC on them.
The chosen query is scored on the cases of all other patients. For LUNA16, the cases are drawn from the training split of~\cite{fmcib} and scored on the test cases of all other patients.
The last column of \cref{tab:supp-fewshot} is the best candidate on the test cases themselves, an upper bound that uses the test labels.
Choosing among about 730 candidates instead of the written list (the concepts present in at least 20 training scans other than normal states, the 18 label names, and the written list) needs 32 cases per class to come within 2 points of the fixed query.

\begin{table}[h]
\caption{Few-shot query choice (AUROC, mean of 1,000 draws, paired difference to the fixed query in parentheses). \method is the mean of three seeds.}
\label{tab:supp-fewshot}
\centering
\footnotesize
\setlength{\tabcolsep}{4pt}
\begin{tabular*}{\textwidth}{@{\extracolsep{\fill}}lrrcccccl@{}}
\toprule
Task & Cases & Positive & Fixed query & $k=1$ & $k=4$ & $k=16$ & $k=32$ & Best on test \\
\midrule
MosMed, COVID-19 & 1,110 & 856 & 95.0 & 94.0 ($-1.1$) & 95.1 ($+0.0$) & 96.1 ($+1.0$) & 96.3 ($+1.3$) & 97.0, ground glass opacity \\
STOIC, COVID-19 & 2,000 & 1,205 & 84.4 & 77.9 ($-6.5$) & 80.5 ($-3.9$) & 83.0 ($-1.3$) & 83.6 ($-0.8$) & 84.4, viral pneumonia \\
STOIC, severe outcome & 1,205 & 301 & 41.3 & 62.3 ($+21.0$) & 65.1 ($+23.9$) & 68.8 ($+27.5$) & 70.5 ($+29.3$) & 73.2, pulmonary edema \\
LUNA16, malignancy & 170 & 82 & 92.4 & 87.9 ($-4.4$) & 90.0 ($-2.4$) & 90.8 ($-1.4$) & 90.8 ($-1.2$) & 92.4, lung mass \\
LUNA25, malignancy & 6,163 & 555 & 74.0 & 71.4 ($-2.6$) & 73.6 ($-0.5$) & 75.5 ($+1.3$) & 76.6 ($+2.1$) & 77.1, mass \\
\bottomrule
\end{tabular*}
\end{table}

\end{document}

%% file: tab_perfinding.tex
\begin{tabular}{@{}lccccc@{\hspace{1.5em}}ccccc@{}}
\toprule
 & \multicolumn{5}{c}{CT-RATE} & \multicolumn{5}{c}{RAD-ChestCT} \\
Finding & \method & Sent. & No st. & Strict & Labels & \method & Sent. & No st. & Strict & Labels \\
\midrule
Pleural effusion & 98.1 & 97.1 & 96.9 & 96.6 & 98.5 & 96.5 & 95.7 & 95.3 & 93.2 & 96.5 \\
Pericardial effusion & 94.8 & 92.8 & 73.6 & 92.0 & 95.1 & 87.5 & 82.0 & 68.0 & 81.5 & 87.6 \\
Cardiomegaly & 94.6 & 93.9 & 94.2 & 87.1 & 94.9 & 91.7 & 90.8 & 91.5 & 83.7 & 91.3 \\
Hiatal hernia & 87.6 & 83.5 & 87.6 & 73.0 & 88.0 & 84.2 & 82.1 & 83.9 & 69.9 & 83.9 \\
Emphysema & 84.9 & 83.6 & 83.7 & 79.2 & 84.8 & 93.4 & 90.3 & 93.9 & 70.3 & 94.6 \\
Bronchiectasis & 87.2 & 79.5 & 86.1 & 76.3 & 87.2 & 85.5 & 83.6 & 81.2 & 74.6 & 85.3 \\
Consolidation & 92.7 & 87.4 & 91.0 & 73.8 & 93.0 & 85.3 & 82.9 & 85.0 & 70.9 & 84.7 \\
Atelectasis & 80.9 & 75.0 & 79.8 & 68.2 & 81.3 & 73.6 & 67.9 & 73.2 & 62.1 & 73.5 \\
Lung opacity & 86.6 & 68.0 & 78.7 & 85.4 & 88.9 & 67.3 & 62.1 & 65.9 & 66.0 & 67.2 \\
Lung nodule & 76.6 & 64.7 & 70.4 & 62.2 & 77.7 & 75.2 & 62.9 & 70.9 & 65.8 & 75.3 \\
Interlobular septal thickening & 89.3 & 84.2 & 89.1 & 79.7 & 90.4 & 86.4 & 80.7 & 86.8 & 76.6 & 86.5 \\
Peribronchial thickening & 84.4 & 81.2 & 83.1 & 76.0 & 84.7 & 74.2 & 63.2 & 73.1 & 61.9 & 74.4 \\
Mosaic attenuation pattern & 91.4 & 90.6 & 91.1 & 67.3 & 91.6 & -- & -- & -- & -- & -- \\
Pulmonary fibrotic sequela & 74.2 & 69.2 & 73.1 & 59.3 & 76.2 & 83.1 & 82.9 & 82.6 & 85.1 & 61.9 \\
Lymphadenopathy & 76.6 & 73.2 & 74.5 & 70.3 & 79.6 & 80.5 & 73.7 & 77.4 & 64.1 & 78.6 \\
Medical material & 89.2 & 88.7 & 88.7 & 85.6 & 96.0 & 76.9 & 73.7 & 75.6 & 74.9 & 82.5 \\
Arterial wall calcification & 94.5 & 92.9 & 94.4 & 87.7 & 95.0 & 83.0 & 82.5 & 83.7 & 73.6 & 83.5 \\
Coronary artery wall calcification & 94.5 & 93.5 & 94.6 & 82.5 & 95.0 & -- & -- & -- & -- & -- \\
\bottomrule
\end{tabular}

%% file: tab_agreement.tex
\begin{tabular}{@{}lrcc@{}}
\toprule
Label & Positive scans & Name alone & Name, location and detail \\
\midrule
Hiatal hernia & 3,426 & 0.98 & 0.98 \\
Emphysema & 4,677 & 0.98 & 0.98 \\
Mosaic attenuation pattern & 1,794 & 0.97 & 0.97 \\
Bronchiectasis & 2,411 & 0.96 & 0.95 \\
Pleural effusion & 2,858 & 0.96 & 0.96 \\
Pericardial effusion & 1,702 & 0.95 & 0.95 \\
Cardiomegaly & 2,586 & 0.94 & 0.94 \\
Atelectasis & 6,170 & 0.91 & 0.92 \\
Interlobular septal thickening & 1,902 & 0.90 & 0.94 \\
Consolidation & 4,234 & 0.87 & 0.96 \\
Lung nodule & 10,974 & 0.87 & 0.80 \\
Medical material & 2,886 & 0.87 & 0.59 \\
Pulmonary fibrotic sequela & 6,447 & 0.86 & 0.91 \\
Peribronchial thickening & 2,496 & 0.84 & 0.87 \\
Lung opacity & 8,846 & 0.82 & 0.81 \\
Arterial wall calcification & 6,701 & 0.43 & 0.84 \\
Lymphadenopathy & 6,115 & 0.29 & 0.36 \\
Coronary artery wall calcification & 5,972 & 0.00 & 0.79 \\
\bottomrule
\end{tabular}